\documentclass[letterpaper, 13pt]{article}
\usepackage{aaai}
\usepackage{times}
\usepackage{helvet}
\usepackage{courier}
\usepackage{amsmath,graphicx,amssymb,subfig, booktabs, relsize}
\usepackage{lettrine}
\usepackage{textcomp}
\begin{document}
%
\title{A Machine Learning Framework for Authorship Identification From Texts}
\author{\bf Rahul Radhakrishnan Iyer$^1$,  {\bf Carolyn Penstein Ros\'{e}$^1$} \\ \\
 $^1$ Language Technologies Institute, School of Computer Science, Carnegie Mellon University, Pittsburgh, PA, USA}

\maketitle
\begin{abstract}
\begin{quote}
Authorship identification is a process in which the author of a text is identified. Most known literary texts can easily be attributed to a certain author because they are, for example, signed. Yet sometimes we find unfinished pieces of work or a whole bunch of manuscripts with a wide variety of possible authors. In order to assess the importance of such a manuscript, it is vital to know who wrote it. In this work, we aim to develop a machine learning framework to effectively determine authorship. We formulate the task as a single-label multi-class text categorization problem and propose a supervised machine learning framework incorporating stylometric features. This task is highly interdisciplinary in that it takes advantage of machine learning, information retrieval, and natural language processing. We present an approach and a model which learns the differences in writing style between $50$ different authors and is able to predict the author of a new text with high accuracy. The accuracy is seen to increase significantly after introducing certain linguistic stylometric features along with text features.\\~\\

\-\hspace{0.5cm} \textbf{\textit{key words ---}} author identification, stylometry, text mining, multi-class classification, information retrieval, natural language processing, machine learning
\end{quote}
\end{abstract}


\noindent \section{Introduction}
\label{sec:intro}
\lettrine{T}he world is evergrowing with texts and it becomes pertinent at times to resolve conflicts of authorship. Most known literary texts can easily be attributed to a certain author because they are, for example, signed. Yet sometimes we find unfinished pieces of work or a whole bunch of manuscripts with a wide variety of possible authors. In order to assess the importance of such a manuscript, it is vital to know who wrote it.

Ways to determine such an authorship has been around for a long time, ever since the days of handwritten manuscripts. Documents then used to be attributed to authors based on the handwriting. But now-a-days, everything is digitalized and this problem is particularly motivated due to applications in the field of forensic analysis. It becomes pertinent to employ some kind of linguistic stylometric analysis to determine the writing style of authors.

\subsection{Related Work}
Authorship identification is not a research area that emerged out of the increased use of internet. It was used for determining which author wrote a chapter or page of a book. Authorship identification research makes use of the structure of text and the words that are used. A subdivision of this is stylometric research in which linguistic characteristics are used to identify the author of a text. Actually, most of the features used for authorship identification are stylometric, especially in literary authorship. In stylometry research, it is generally accepted that authors have unconscious writing habits ~\cite{chaski1997wrote,anderson2001identifying,corney2003analysing}. These habits become evident in for example their use of words and grammar. The more unconscious a process is, the less controllable it is. Therefore, words and grammar could be a reliable indication of the author. These individual differences in use of language is referred to as idiolect. The unconscious use of syntax gives rise to the opportunity to perform author identification based on stylometric features. 

A commonly used stylometric feature is based on n-grams of characters. Experiments that use n-grams of characters have shown to be successful in determining the authorship of texts ~\cite{clement2003ngram,corney2003analysing}. Also, structural information is relevant for determining authorship, successful classifications are reported when using bigram of syntactic labels ~\cite{hirst2007bigrams}. 

Several aspects can influence the performance of this task, such as language of the messages used, the length of these messages, the number of authors and messages, the types of features and the classification method. The number of features is most often varied to determine the influence of certain types of features. Corney et al. ~\cite{anderson2001identifying} indicate that the most successful features are function words and character n-grams. Their tests on non-email data showed that function words gave good results independent of topic, while the character n-gram seemed to depend on topic. They performed several experiments, each with a different set of stylometric features.

De Vel et al. ~\cite{de2001mining} also executed experiments with email messages. They used $156$ messages from three native English authors. Each author contributed e-mails on three topics and the classification was performed using $170$ stylistic features and $21$ features describing the structure of the mail.

McCombe, on her master's thesis, executed experiments to determine which features can successfully be used for authorship identification. The author performed tests using word unigrams and showed that the results using this method were promising. But no method she used was successful in classification based on word bigrams. Hirst and Feiguina ~\cite{hirst2007bigrams} used tag bigrams to discriminate between the work of Anne and Charlotte with three experiments using tag bigrams.

Zheng et al. ~\cite{zheng2006framework} executed an experiment, which revealed that the performance increases when the number of authors drop. This holds for several machine learning algorithms for English as well as Chinese. The research of Houvardas and Stamatatos ~\cite{houvardas2006n} and the research of Van Der Knaap and Grootjen ~\cite{van2007author} show that successful results can be obtained when many texts and more than $25$ authors are used. Recently, several approaches involving natural language processing \cite{iyer2019event,iyer2019unsupervised,iyer2019heterogeneous,iyer2017detecting,iyer2017recomob}, machine learning \cite{li2016joint,iyer2016content,honke2018photorealistic}, deep learning \cite{iyer2018transparency,li2018object} and numerical optimizations \cite{radhakrishnan2016multiple,iyer2012optimal,qian2014parallel,gupta2016analysis,radhakrishnan2018new} have also been used in the visual and language domains. \\

In this work, we attempt to introduce some new stylometric features and investigate the use of POS (part-of-speech) tags into our feature space. Using POS sequences or POS Bigrams has not been explored in previous works. We think it could be very relevant to this problem because it could happen that each author has a unique sequence of POS tags of words that he/she uses when writing his/her texts. This has been shown true to some extent, owing to the increase in performance after incorporating POS tags and POS Bigrams into the feature space. Thus, our contribution to the literature is to explore such new feature spaces, and to combine them with stylometric features to increase the performance.
\subsection{Paper Organization} 
The paper is organized as follows. Section (\ref{sec:data_coll}) discusses the dataset being used in the experiments. Data preparation is discussed in section (\ref{sec:data_prep}). This section talks about certain aspects of the dataset, including its division for the experiments performed. We formulate the problem and perform exploratory data analysis in sections (\ref{sec:probform}) and (\ref{sec:data_anal}) respectively. Section (\ref{sec:experiments}) presents a baseline performance, and error analysis including adding linguistic stylometric features to the dataset, and section (\ref{sec:optimization}) talks about feature selection and parameter tuning. The experimental results obtained  and their implications are discussed in section (\ref{sec:results}). We present a discussion of our work and draw conclusions in section (\ref{sec:discussion}), along with few limitations of our work in section (\ref{sec:limits}) and finally explore possibilities of future work in the last section (\ref{sec:concl}) of the paper.

\section{Data Collection}
\label{sec:data_coll}
The dataset used for this experiment is a subset of the popular and well-established RCV1 (Reuters Corpus Volume 1) dataset, used as a benchmark for research in information retrieval, called the \textit{Reuter-50-50 dataset}. RCV1 is basically a corpus of newswire stories made public by Reuters Ltd. 

In 2000, Reuters Ltd. made available a large collection of Reuters news stories for use in research and development of natural language processing, information retrieval, and machine learning algorithms. RCV1 is an archive of over $800,000$ manually categorized newswire stories. It was intended to consist of all and only English language stories produced by Reuters journalists between August $20$, $1996$ and August $19$, $1997$.

This subset that we are using, \textit{Reuters-50-50}, has been used in many author identification experiments. This dataset consists of a total of $5000$ instances, with each instance being a news story written by an author. There are a total of $50$ authors (the top $50$ authors, in terms of total size of articles, were selected from the RCV1 dataset), each having $100$ texts/news stories associated with them, thus making this a single-label mutli-class classification problem.

This is a very balanced dataset because there are an equal number of instances for each class, each class being an author. The fifty different authors are shown in Table ~\ref{authors}.

\begin{table}
 \vspace{2ex}
 \resizebox{\columnwidth}{!}{
 \begin{tabular}{l | l | l}
 \toprule
 \multicolumn{3}{c}{\textbf{Authors}}\\

 \midrule

 Aaron Pressman & Alan Crosby & Alexander Smith\\
 Benjamin Kang Lim & Bernard Hickey & Brad Dorfman\\
 Darren Schuettler & David Lawder & Edna Fernandes\\
 Eric Auchard & Fumiko Fujisaki & Graham Earnshaw\\
 Heather Scoffield & Jane Macartney & Jan Lopatka\\
 Jim Gilchrist & Joe Ortiz & John Mastrini\\
 Jonathan Birt & Jo Winterbottom & Karl Penhaul\\
 Keith Weir & Kevin Drawbaugh & Kevin Morrison\\
 Kirstin Ridley & Kourosh Karimkhany & Lydia Zajc\\
 Lynne O'Donnell & Lynnley Browning & Marcel Michelson\\
 Mark Bendeich & Martin Wolk & Matthew Bunce\\
 Michael Connor & Mure Dickie & Nick Louth\\
 Patricia Commins & Peter Humphrey & Pierre Tran\\
 Robin Sidel & Roger Fillion & Samuel Perry\\
 Sarah Davison & Scott Hillis & Simon Cowell\\
 Tan Ee Lyn & Therese Poletti & Tim Farrand\\
 Todd Nissen & William Kazer & \\
 
 \bottomrule
 \end{tabular}
 }
 \centering
 \caption{List of all the $50$ authors in the dataset}
 \label{authors}
 \end{table}

\section{Data Preparation}
\label{sec:data_prep}
The dataset at hand, as mentioned earlier, is very balanced. Keeping that in mind, the dataset was divided into \textit{three} sets:
\begin{itemize}
	\item \textit{Development set}
	\item \textit{Cross-Validation set}
	\item \textit{Final holdout test set}
\end{itemize}

The \textit{development set} is used for all the feature engineering procedures: adding features, error analysis, feature selection etc. Error analysis and feature selection are done by building a model on the \textit{cross-validation} set and then testing on the \textit{development set}. The \textit{cross-validation} set is used for evaluating the performance of the model. We make this split between \textit{development} and \textit{cross-validation} sets to try to avoid overfitting of the model. If we engineer features, analyze the errors and evaluate the performance all on one set, then our analysis will tend to ``favor'' that set and we cannot hope for it to generalize to unseen data. This is because we are seeing \textit{all} the data before we make our decision, in some sense. Thus, we split our data into \textit{development} and \textit{cross-validation}. The \textit{holdout test set} is used to test the final performance of the model, after optimizing it using the \textit{development} and \textit{cross-validation} sets.

There were a total of $5000$ instances in the dataset as pointed out earlier ($100$ texts for each of the $50$ authors). Since, there is an equal distribution of class values (authors) in the dataset, we have split the dataset such that the uniformity is maintained in the subsets too. Out of the $5000$, $1000$ instances were assigned to the \textit{development} set, $3500$ instances for the \textit{cross-validation} and finally $500$ instances for the \textit{holdout test set}. Each of these subsets also had an equal distribution of class values. The distribution is given in Table ~\ref{dist}.

\begin{table}
 \vspace{2ex}
 \resizebox{\columnwidth}{!}{
 \begin{tabular}{l | c | c | c}
 \toprule
 \textbf{Dataset} & \textbf{\# instances} & \textbf{\# authors} & \textbf{\# texts/author}\\

 \midrule
 \textit{development} & 1000 & 50 & 20\\
 \textit{cross-validation} & 3500 & 50 & 70\\
 \textit{holdout test} & 500 & 50 & 10\\
 
 \bottomrule
 \end{tabular}
 }
 \centering
 \caption{Distribution of instances in the different datasets}
 \label{dist}
 \end{table}

We used LightSide for our experiments. 
We extracted \textit{unigram} features from the dataset without any punctuation. These set of features were then used on the \textit{development} set for choosing the best performing algorithm. More features were added during error analysis.

\section{Problem Formulation}
\label{sec:probform}
In this section, we formulate the problem formally. We assume a collection $D = [d_1, d_2, \ldots, d_n]$ of text documents and $A = [a_1, a_2, \ldots, a_m]$ of relevant authors. The task at hand is quite simple: for any given document $d_i \in D$, we are required to predict the most suitable author $a_j \in A$ for that text and by most suitable, we mean the author who would have most likely written that document. This can also be thought in a probabilistic way: given a document $d_i \in D$, we predict the most likely author $a_j \in A$ such that $P(a_j|d_i)$ is maximized i.e. $\displaystyle\mathop{\mathlarger{\mathlarger{\forall}}}_{a_k \neq a_j} a_k \in A, P(a_j|d_i) > P(a_k|d_i)$.

\section{Data Exploration} 
\label{sec:data_anal}
Exploratory data analysis was performed using the \textit{development} set. This included manual examination of the dataset as to what kind of features can be used etc. In addition, feature selection was done using the \textit{development} set: the model was built on the \textit{cross-validation} set and tested on the \textit{development} set. We then choose that number of features to be selected which resulted in the best performance of the model. This is explained in greater detail in section (\ref{sec:optimization}). 

As mentioned before, we extracted unigram features (with no punctuation) from the \textit{development} (the set has only two columns: class and text) using Lightside. There were a total of $6488$ unigram features. To gain an initial estimate of what kind of performance can be expected, several algorithms were run on the \textit{development} set. In particular, algorithms like LibLINEAR SVM, SMO, Logistic Regression, Na\"ive Bayes and J$48$ Decision Tree, were chosen and run. Each of these algorithms was run in Lightside to predict the authors, using a $10$-fold cross-validation on the \textit{development} set. All the algorithms were run with their default settings. Their performance values are listed in Table ~\ref{dev_algs}.

\begin{table}
 \vspace{2ex}
 \begin{tabular}{l | c | c}
 \toprule
 \textbf{Algorithm} & \textbf{\% Correct} & \textbf{Kappa}\\

 \midrule
 LibLINEAR SVM & \textbf{83.2} & \textbf{0.8286}\\
 SMO & 79.1 & 0.7867\\
 Logistic Regression & 82.2 & 0.8188\\
 Na\"ive Bayes & 63.7 & 0.6296\\
 J$48$ Decision Tree & 55 & 0.5408\\
 
 \bottomrule
 \end{tabular}
 \centering
 \caption{Results of using $10$-fold cross-validation on the \textit{development} set. All settings are default unless otherwise indicated. The metrics used to evaluate the results are: percent correct and Cohen's kappa.}
 \label{dev_algs}
 \end{table}

We hypothesized that LibLINEAR SVM would provide the best results, based on observing its performance in several classification tasks. As can be seen from the above table, we see that the best-performing algorithm is LibLINEAR SVM, which is a linear model. Since both SVM and Logistic Regression, both being linear models, have performed very well on the data set, it is safe to assume that the dataset is \textit{linearly separable}. Thus, we see that the best suited model for this dataset is \textit{linear} and \textit{weight-based}. We see that the values of kappa for the algorithms is high, which suggests that there is much agreement without chance, which is a good thing. One can also observe that Na\"ive Bayes performs poorly, implying that its assumption of conditional independence of the features given the class value is not that accurate.

There are several advantages with linear models such as SVM. Firstly, it is a stable algorithm and is protected to some extent against overfitting, and hence does not require ensemble techniques like bagging, boosting, stacking to make it more robust. In addition, SVM, being a linear model, has the added capability of ignoring noisy attributes because linear models achieve results through a focus on achieving higher accuracy (the process being similar to that of a Decision Tree). So, this is a good robust algorithm to work with.

We did not have to do any \textit{Data Cleansing}, such as removing outliers from the dataset. This is because it was observed that not a lot of instances were commonly misclassified by all the algorithms mentioned above. This suggested that there were not many outliers and thus such a cleansing procedure was not necessary. In addition, after going through the dataset, we did not observe subpopulations. This was also determined by noticing that the result obtained was not that inflated: a very similar result was obtained by performing cross-validation on the \textit{development} set, cross-validation on the \textit{cross-validation} set and even training a model on the \textit{cross-validation} set and testing on the \textit{development} set. Thus, we can avoid taking measures against subpopulations in our data such as \textit{feature splitting}, creating domains. Also, since there is no skew in our dataset and all the classes have an equal distribution, we did not need to use \textit{Cost-sensitive classification} for our experiments.\\

From everything that is said above, we choose our final algorithm to work with as the LibLINEAR SVM.

\section{Experiments}
\label{sec:experiments}
In this section, we discuss the experiments that we have conducted: baseline performance, error analysis and adding new features.

\subsection{Baseline Performance}
Baseline performance was obtained by running the LibLINEAR SVM algorithm in Lightside to predict the authors, using a $10$-fold cross-validation on the \textit{cross-validation} set. The algorithm was run with its default settings. In addition, only the unigram features were extracted from the \textit{cross-validation} set, without punctuation, for the baseline. A total of $14300$ unigram features were extracted. This gave an accuracy of $88.83$\% and a kappa statistic of $0.886$\footnote{It may be noted that the baseline performance presented here does not match with that presented during the first phase in September. That is because we had not used all the unigram features, and only considered the first $2000$ or so for the experiments. Considering all of them has given a much better performance.}. There is a lot of scope for improvement and we detail below the analysis taken to obtain better performance.

\subsection{Error Analysis}
To further refine the model, an error analysis was performed. To do this, the model was first trained on the \textit{cross-validation} set and then tested on the \textit{development} set, having extracted only the unigram features. There were a total of $14300$ unigram features. The performance obtained was: accuracy of $79.3$\% and a kappa statistic of $0.7888$. In Lightside, we go to see which error cells in the confusion matrix has the most errors, which can be corrected.  Since there are $50$ different class values, the dimension of the confusion matrix is $50 \times 50$. The cells with the maximum error was found to be $\left( 35, 45 \right)$, which had $7$ misclassified instances. 

Probing in to see what kind of instances were misclassified, we found something that we were expecting. It was very interesting to note that different authors conveyed similar messages differently. In essence, the news that the text represents could talk about the same news but the writing style differed. We observed that many of the authors used two words together in most contexts. For example, one problematic feature that we observed was the word \textit{official}. It had a high horizontal difference, feature weight and frequency. Some authors like Nick Louth has used the word \textit{official} in only certain contexts: like \textit{government official}. But others have used it in several other contexts like: \textit{it is official, on official business, official statement} etc. Some of the misclassified instances were like this. To correct this, if we were to include context information to the features, then if \textit{government official} appears, the model will not make a mistake to assign that text to Nick Louth. To account for this, we include bigram features. Including bigram features will take care of this problem because we see two words at a time and this gives the words more context. In addition to this, we can also look at the POS bigrams as that has also some pattern for different authors.

In addition to this, we also observed that different authors had their own way of using POS tags of the words. Thus, we came to the conclusion that combining POS tags into the feature space could make the model learn to distinguish the authors better. Thus, the textual features that we considered to be included in the current feature space, which included only unigrams: bigrams, POS bigrams and word/POS pairs. These would give a very good representation of an author's writing: the kinds of words one tends to use together along with POS tags etc. are all captured by these features.

To further our performance and get a more accurate representation of an author's style, we looked at some additional meta features: linguistic stylometric features. From each of the texts, we did:
\begin{itemize}
	\item Phraseology Analysis: Features about the phrases etc. used
	\item Punctuation Analysis (per 1000 tokens): Features about the punctuations
	\item Lexical Usage Analysis (per 1000 tokens): Features about the usage of certain words
\end{itemize}

Several features were computed during each of these analysis. These features are listed in Table ~\ref{stylometric_features_1}, ~\ref{stylometric_features_2} and ~\ref{stylometric_features_3}.

\begin{table}
 \vspace{2ex}
 \resizebox{\columnwidth}{!}{
 \begin{tabular}{l | c | l}
 \toprule
 \multicolumn{3}{c}{\textbf{Phraseology Analysis}}\\
 \textbf{Name} & \textbf{Type} & \textbf{Explanation}\\
 \midrule
 Lexical diversity & numeric & Lexical Diversity\\
 Mean Word Length  & numeric & Mean Word Length\\
 Mean Sentence Length & numeric & Mean Sentence Length\\
 STDEV Sentence Length & numeric & STDEV Sentence Length\\
 Mean paragraph Length & numeric & Mean paragraph Length\\
 Document Length & numeric & Document Length\\
 
 \bottomrule
 \end{tabular}
 }
 \centering
 \caption{Features from Phraseology Analysis of text. Here, lexical diversity refers to the ratio of total number of words to the number of different unique word stems}
 \label{stylometric_features_1}
 \end{table}

\begin{table}
 \vspace{2ex}
 \resizebox{\columnwidth}{!}{
 \begin{tabular}{l | c | l}
 \toprule
 \multicolumn{3}{c}{\textbf{Punctuation Analysis}}\\
 \textbf{Name} & \textbf{Type} & \textbf{Explanation}\\
 \midrule
 Commas & numeric & \# Commas\\
 Semicolons  & numeric & \# Semicolons\\
 Quotations & numeric & \# Quotations\\
 Exclamations & numeric & \# Exclamations\\
 Colons & numeric & \# Colons\\
 Hyphens & numeric & \# Hyphens\\
 Double Hyphens & numeric & \# Double Hyphens\\
 
 \bottomrule
 \end{tabular}
 }
 \centering
 \caption{Features from Punctuation Analysis of text}
 \label{stylometric_features_2}
 \end{table}

\begin{table}
 \vspace{2ex}
 \begin{tabular}{l | c | l}
 \toprule
 \multicolumn{3}{c}{\textbf{Lexical Usage Analysis}}\\
 \textbf{Name} & \textbf{Type} & \textbf{Explanation}\\
 \midrule
 and & numeric & \# and\\
 but  & numeric & \# but\\
 however & numeric & \# however\\
 if & numeric & \# if\\
 that & numeric & \# that\\
 more & numeric & \# more\\
 must & numeric & \# must\\
 might & numeric & \# might\\
 this & numeric & \# this\\
 very & numeric & \# very\\
 
 \bottomrule
 \end{tabular}
 \centering
 \caption{Features from Lexical Usage Analysis of text}
 \label{stylometric_features_3}
 \end{table}

There are a total of $23$ stylometric features that we had added to the feature space. These capture some very certain writing styles and practices. we believed that the writing style could be captured much better with this new feature space that includes textual and stylometric features. We extracted these features because we believed that just textual features will not be sufficient always to determine authorship. We may, at times, need to look at certain non-textual parts too. This is what motivated us to look for stylometric features.

\begin{figure}
\centering
\includegraphics[scale=0.7]{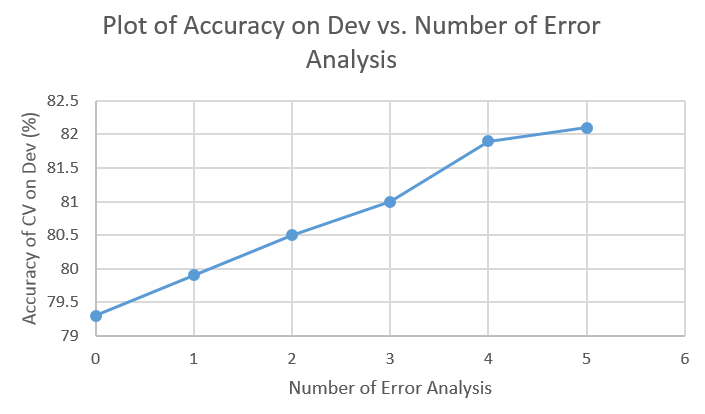}
\caption{Progress of Error Analysis throughout the experiment}
\label{error_analysis}
\end{figure}

Now that the motivation has been provided for these features, these were then extracted and a new feature space was formed. In order to test the performance of this new space, we ran a $10$-fold cross validation on the \textit{cross-validation} set using the LibLINEAR SVM algorithm, in the new feature space. The new feature space, that included unigrams, bigrams, POS bigrams, word/POS pairs and stylometric features, had a total of $82549$ features. The result obtained was: accuracy of $91.29$\% and a kappa statistic of $0.9111$. This is a \textbf{highly significant improvement} over the baseline, which had an accuracy of $88.83$\% and a kappa statistic of $0.886$. The significance test was performed on Lightside using the paired t-test (p = 0**, t = $-6.932$). Thus, the error analysis was a success and we have obtained an increased performance. This suggests that the stylometric features, as predicted to make a difference making the algorithm learn better. A test was also performed on the \textit{development} set, after training the model on the \textit{cross-validation} set. This performance was: accuracy of $81.9$\% and a kappa statistic of $0.8153$, which is much better than the unigram case (before error analysis), which had an accuracy of $79.3$\% and a kappa statistic of $0.7888$. The graph in Fig. ~\ref{error_analysis} shows  the progress of error analysis throughout the experiment: the plot is the accuracy obtained when the model is trained on the \textit{cross-validation} set and tested on the \textit{development} set.

\section{Optimization}
\label{sec:optimization}
In this section, we discuss the optimization strategies used to tune the algorithm LibLINEAR SVM further. Firstly, we do a feature selection to pick the most predictive subset of features from the large set of features. After that we tune certain parameters of the algorithm.
\subsection{Feature Selection}
\begin{figure}
\centering
\includegraphics[scale=0.7]{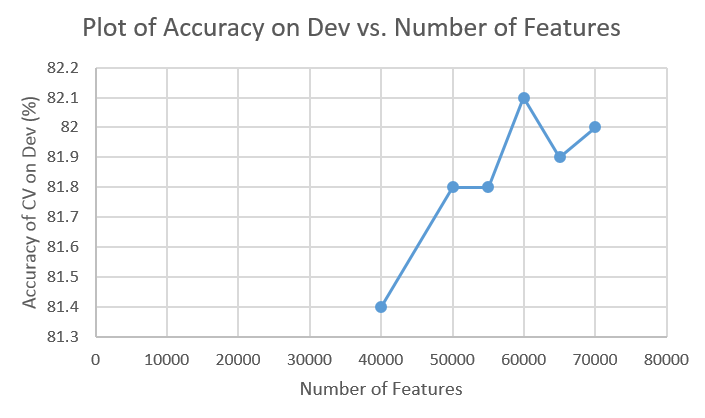}
\caption{Performance of the choice of number of features to select}
\label{feature_selection}
\end{figure}

Here, we want to select the most predictive subset of features from the whole feature space. This is so that we can avoid unnecessary and noisy features. We pick the model after the error analysis. This model has a total of $82549$ features, which is quite a lot. This model is trained on the \textit{cross-validation} set. We aim to find a suitable subset of features that reduces this value. Feature selection involved using \textit{AttributeSelectedClassifier} with \textit{ChiSquaredAttributeEval} and \textit{Ranker} selecting the top \textit{n} attributes, where \textit{n} is the number of attributes we want to choose. We try different numbers like: $40000$, $50000$, $55000$, $60000$, $65000$ and $70000$. To pick the best number of features, we have to use the \textit{development} set as the test set. In essence, the model that is trained on the \textit{cross-validation} set is tested on the \textit{development} set for each of these number of features. We then pick that number of features which gives us the best performance. We find that $60000$ features in fact gives us the best performance. Graph ~\ref{feature_selection} shows the performance for each of the number of features selected. This optimized model gives a significant (highly) improvement over the baseline again (p = 0**, t = 0.632). A $10$-fold cross-validation performed on the \textit{cross-validation} set using this optimized model gives an accuracy of $91.3$\% and a kappa statistic of $0.9121$. This is not a significant improvement over the previous model before the feature selection. Nonetheless, it is still an improvement.

\subsection{Parameter Tuning}
Here, we want to tune the parameters of the LibLINEAR SVM algorithm. The objective is to find the parametric value that give us the best performance for the algorithm on our dataset. The parameters that were available to tune in this algorithm were: bias term, cost parameter, tolerance value and the weights. None of cost parameter, tolerance value or the weights had any effect on the accuracy or kappa (though the confusion matrices changed a bit). We had tried atleast $4$ different values for each of them. Since, there was no other parameter to tune, we had to settle with tuning the \textit{bias} term. The default value for bias is $1$.
We did the tuning manually and did not use \textit{CVParameterSelection} to achieve this. We chose $3$ different values for our bias term apart from the default (default is $1$): $3$, $4$, $5$. To tune the model, the feature-selection-tuned model was used, i.e the new model with $60000$ features was used. Each of these parameter settings was tested using $10$-fold cross validation on the \textit{cross-validation} set. This model is built on the \textit{cross-validation} set and then tested on itself using a $10$-fold cross-validation for each of the parameter settings. The results for each of these is given in Table ~\ref{tuning}

\begin{table}
 \vspace{2ex}
 \begin{tabular}{l | c | c}
 \toprule
 \textbf{Bias Term} & \textbf{Accuracy \%} & \textbf{Kappa}\\
 \midrule
 1 & 91.3 & 0.9121\\
 3 & 91.2 & 0.9102\\
 4 & 91.17 & 0.9006 \\
 5 & 91.19 & 0.9009\\
 
 \bottomrule
 \end{tabular}
 \centering
 \caption{Performances for the different parameter setting choices}
 \label{tuning}
 \end{table}

As can be seen from the above results, the best performing setting for bias is the default setting itself, which is $1$. Thus, we don't tune any parameter for the LibLINEAR SVM and proceed with this as the final model for testing on the \textit{holdout test} set.
 
\section{Results}
\label{sec:results}
Now, we have finally built a robust model that we hope will generalize to unseen data. The final model after feature selection is chosen to test on the \textit{holdout test} set. Here, we show a comparison of the performance of the optimized model and the baseline model on the \textit{holdout test} set. Firstly, the baseline model is trained on the \textit{cross-validation} set with the baseline features (unigram), and then tested on the \textit{holdout test} set. The performance obtained: accuracy of $79.8$\% and a kappa statistic of $0.7939$. After that, we train the optimized model on the same \textit{cross-validation} set with the new feature space, and then tested on the \textit{holdout test} set. The performance obtained: accuracy of $81.6$\% and a kappa statistic of $0.8122$. As can be seen, the performance of the optimized model is definitely better. But upon doing a significance test, we see that the improvement is not significant (p = $0.106$, t = $-1.619$).

\section{Discussion and Conclusions}
\label{sec:discussion}
This paper has explored machine learning tools and techniques to predict the authors of texts. This has included data preparation, baseline performance observation, error analysis and optimization (tuning and feature selection). Over $80$\% of the instances in the \textit{holdout test} set were correctly predicted by the optimized LibLINEAR SVM. The error analysis was a complete success and the improvement achieved was \textit{highly significant}.
Most of the previous work done have tried to explore stylistic and linguistic features to accomplish this task. But as far as we've seen, we have not seen any of the previous work try our combination of features: clubbing stylometric meta features with textual features like bigrams, POS bigrams and word/POS pairs. This is our contribution to this field. 
The accuracy increases significantly after introducing stylometric features, which guarantees our premise that such features combined with bigrams is more predictive of an author's writing style than just unigrams. This result is consistent with the previous research. This field is a very exciting one, and this work particularly could be of immense use to the forensics team, when they would need to solve disputes in authorship of texts.

\section{Limitations}
There are some limitations to the work presented here. Firstly, the analysis done, though robust is not good enough. A lot more evaluations and a lot more metrics need to be considered: like precision, recall, F-measure, discounted cumulative gain etc. From the above results, though we can say that the result generalized to some extent (owing to the performance on the \textit{holdout test} set), nevertheless a more comprehensive evaluation scheme has to be followed. Another limitation is that this work only considers texts in English. There are authorship disputes in other language texts as well. Also, the dataset used here is very restricted to news topics. Finally, the approach used here may not generalize to very short texts, such as determining the author of microblog posts such as tweets. It would be interesting to see what kind of features would solve authorship disputes in very short texts. Also, the type of stylometric features used could be improved.
\label{sec:limits}


\section{Future work}
\label{sec:concl}
There are several ways this work could be extended. Firstly, richer and more diverse dataset could be used to see how the classifier performs (if it's as good as this one). Secondly, including non-english texts could be included into our dataset. This could be challenging for several reasons. The kind of linguistic features that would work for English may not work for Turkish or Chinese, because the linguistic structure of sentences etc. is completely different in the two languages. Thirdly, the work could be made general by considering short texts and long texts in our dataset. Our current model may not be able to handle microblog posts such as facebook statuses or tweets. But since the world is becoming digital day by day, it may become pertinent to classify such texts too. Lastly, better stylometric features could be used.
\newpage
\bibliography{refs}
\bibliographystyle{aaai}
\end{document}